\documentclass[letterpaper, 10 pt, conference]{ieeeconf}  

\IEEEoverridecommandlockouts                              

\usepackage{graphicx}
\usepackage{subcaption}

\usepackage{amsmath} 
\usepackage{amssymb}  
\usepackage{cite}
\usepackage{hyperref}
\usepackage{xcolor}

\usepackage{cleveref}
\Crefname{figure}{Fig.}{Figs.}
\crefname{figure}{Fig.}{Figs.}
\usepackage{algorithm}
\usepackage[noend]{algorithmic}

\usepackage[font=footnotesize,labelsep=period]{caption}

\title{\LARGE \bf
SubCDM: Collective Decision-Making with a Swarm Subset
}

\author{ Samratul Fuady$^{1}$, Danesh Tarapore$^{1}$, and Mohammad D. Soorati$^{1}$
\thanks{This work was supported by the Indonesia Endowment Fund for Education (LPDP).}
\thanks{$^{1}$School of Electronics and Computer Science, University of Southampton, Southampton, SO17 1BJ, UK.
        {\tt\small \{s.fuady, d.s.tarapore, m.soorati\}@soton.ac.uk}}%
}

\begin{document}

\maketitle
\thispagestyle{empty}
\pagestyle{empty}

\begin{abstract}
Collective decision-making is a key function of autonomous robot swarms, enabling them to reach a consensus on actions based on environmental features. Existing strategies require the participation of all robots in the decision-making process, which is resource-intensive and prevents the swarm from allocating the robots to any other tasks. We propose Subset-Based Collective Decision-Making (SubCDM), which enables decisions using only a swarm subset. The construction of the subset is dynamic and decentralized, relying solely on local information. Our method allows the swarm to adaptively determine the size of the subset for accurate decision-making, depending on the difficulty of reaching a consensus. Simulation results using one hundred robots show that our approach achieves accuracy comparable to using the entire swarm while reducing the number of robots required to perform collective decision-making, making it a resource-efficient solution for collective decision-making in swarm robotics.
\end{abstract}

\section{Introduction}

Swarm robotics is a rapidly growing area of research, gaining significant attention due to its broad potential applications across various fields \cite{dorigo2021swarm}. In these systems, large groups of simple robots collaborate through local interactions and self-organization to achieve complex tasks. 
Environmental monitoring \cite{albani2017field} and search-and-rescue missions \cite{horyna2023decentralized} are a few examples where swarm robotics holds great promise. 
A key requirement of swarm robotics is collective decision-making, which enables the swarm to exhibit coordinated, intelligent behavior without any centralized control. This is also known as the best-of-\textit{n} problem \cite{valentini2017best}, where the swarm must choose the best option from a set of $n$ alternatives. It simulates real-world applications such as resource discovery, pollutant detection, and object identification, showing how swarms perceive and interact with their environment through decentralized cooperation.

Optimizing the efficiency of collective decision-making in swarm robotics remains challenging, as deploying large numbers of robots can be costly. Efficiency in swarm systems can be improved in several ways, such as optimizing robot exploration and movement \cite{dimidov2016random}, reducing communication range \cite{talamali2021when}, minimizing network connectivity requirements and decision-making time \cite{reina2024speed}, or decreasing sensing frequency \cite{aust2022hidden}. Despite these efforts, efficiency in terms of reducing the number of robots involved in the decision-making process remains underexplored. Current approaches still rely on the full participation of the swarm to reach a collective decision.

In this work, we explore the potential of using only a subset of the swarm rather than the whole group. We argue that, for simpler tasks, collective decision-making can be achieved with a subset of robots within a swarm, thereby conserving energy and allowing the remaining robots to focus on other tasks if available. For example, in large-scale wildfire monitoring, a subset of drones can handle collective decision-making for detecting fire risks, while others focus on tasks like tracking smoke, assessing damage, or maintaining communication, improving efficiency by conserving resources for decision-making. We propose SubCDM, a method that allows the swarm to autonomously form the decision-making subset and determine its size to maintain accurate decisions, enabling adaptation to varying task difficulty.
We show that our proposed method reduces the number of active robots in collective decision-making while maintaining accuracy similar to using all robots, demonstrated through the ARGoS simulator \cite{pinciroli2012argos} with a swarm of 100 robots.

\section{Related Work}

Various strategies have been developed for achieving collective decision-making in robot swarms (e.g., \cite{valentini2014self,valentini2016collective,ebert2020bayes, kaiser2024learning}). Direct Modulation of Voter-based Decisions (DMVD) \cite{valentini2014self} and Direct Modulation of Majority-based Decisions (DMMD) \cite{valentini2016collective} are the two common strategies that utilize positive feedback modulation and apply voter model and majority rule, respectively. In these methods, robots only share opinions about dominant environmental features. The Direct Comparison (DC) strategy \cite{valentini2016collective} extends these methods by incorporating quality estimates alongside opinions. 
Other variants of collective decision-making methods apply Bayesian inference \cite{ebert2020bayes} and evolved artificial neural networks \cite{kaiser2024learning}.
These methods aim to enhance performance by improving both the speed and accuracy of collective decision-making, though some come with costs such as increased communication bandwidth and more advanced computation.
Additionally, research has explored more complex problems, such as dynamic environments \cite{soorati2019plasticity}, communication limitations \cite{kelly2022collective} and malicious robots \cite{reina2023cross}, further enhancing the robustness and reliability of swarm decision-making. 

The aforementioned collective decision-making strategies require the involvement of the entire swarm of robots, which can be costly. 
Furthermore, swarm performance does not always increase linearly with the addition of more robots \cite{hamann2020guerrilla}. In some cases, performance may plateau as the swarm size increases, or even degrade beyond a certain threshold \cite{kuckling2024do}. Therefore, involving a large number of robots in collective decision-making can be inefficient, particularly in large-scale swarms.
We propose a novel framework that uses self-organized subsets of robots for collective decision-making. This approach enables the swarm to autonomously determine the subset size for accurate and efficient decisions, freeing other robots to perform different tasks and maximizing swarm efficiency.

\begin{figure}[tbp]
    \centering
    \includegraphics[width=0.85\linewidth, trim=10pt 10pt 10pt 10pt, clip]{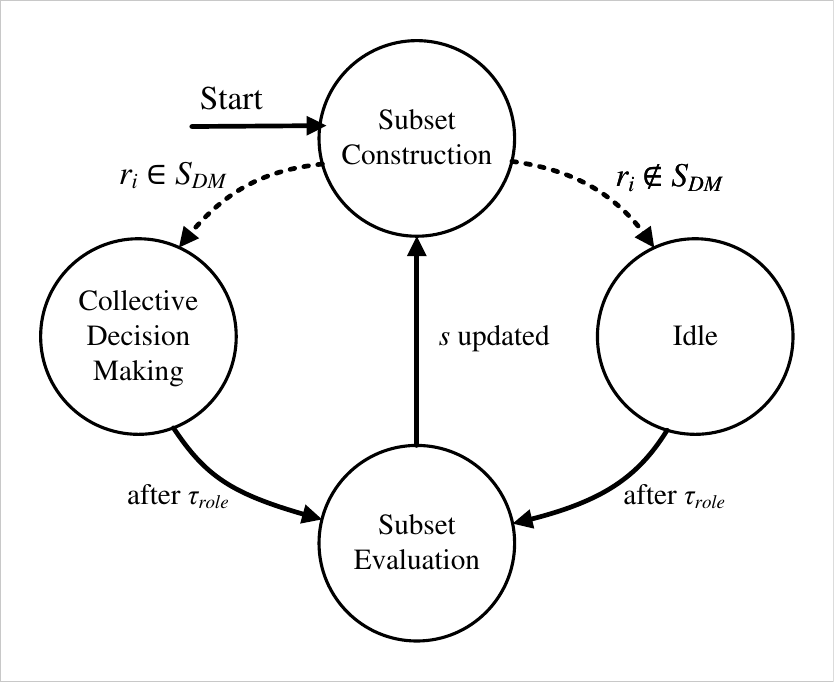}
    \caption{State diagram of SubCDM. Robot $r_i$ either joins the decision-making subset $S_{DM}$ or remains idle based on subset size parameter $s$. Once assigned a role, the robot stays in that role for a randomly sampled duration $\tau_{role}$.}
    \vspace{-10pt}
    \label{fig:state_diagram}
\end{figure}

\section{Method} \label{sec_Method}
We address a best-of-\textit{n} collective decision-making problem ($n=2$) where a swarm of robots must collectively choose between two environmental features. The environment consists of black and white tiles, and the swarm's goal is to determine which color is more dominant (similar to the main body of work in collective decision-making, e.g., \cite{valentini2016collective}). Each robot forms an opinion, denoting 0 if it believes black is more dominant, and 1 otherwise. Unlike conventional collective decision-making strategies that involve the entire swarm, SubCDM utilizes only a subset of robots within the swarm. The process operates through three core phases: subset construction, collective decision-making, and subset evaluation, as shown in \cref{fig:state_diagram}. 

In the subset construction phase, robots self-organize to form the decision-making subset ($S_{DM}$), based on subset parameter $s$, which governs the size of $S_{DM}$. This process is dynamic and asynchronous, enabling robots to adjust their roles based on the situation. Robots in $S_{DM}$ actively participate in decision-making, exploring the environment and sharing information to reach consensus, while other robots remain idle or ready for other tasks. Each robot stays in its assigned subset for duration $\tau_{role}$, allowing the robot to have sufficient time to make a significant contribution. In our experiments, the duration ($\tau_{role}$) is drawn from an exponential random distribution with a mean of 20 seconds. A decision is considered complete when a sufficient percentage of $S_{DM}$ converges on a single opinion. The decision-making subset evaluation phase determines whether the expansion of the subset is needed based on convergence within $S_{DM}$. If $S_{DM}$ fails to converge, $s$ is incremented, and the robots reorganize to form a larger subset, enabling the swarm to scale its resources based on the decision-making task's needs.

We introduce two strategies for implementing SubCDM: leader-based and distributed. Both use only local information to form a subset that can achieve decision accuracy comparable to the entire swarm, using fewer participating robots. The leader-based strategy constructs the subset based on structural connections from the leader, while the distributed strategy does not require any structure to form the subset. The code is available online\footnote{\url{https://github.com/SooratiLab/subset_CDM.git}}.

\subsection{Leader-Based Strategy}

\emph{Subset Construction:}
In this strategy, one robot is designated as the leader, either randomly chosen at the beginning of the run or through a distributed leader election process, e.g. using the method in \cite{vasudevan2004design}; both methods are used in different experiments in this paper. The recruitment of robots for the decision-making subset is based on their geodesic distance from the leader robot, meaning global positioning is not required. Geodesic positioning, often referred to as hop counts, has been previously employed in swarms, e.g. for measuring environmental features \cite{soorati2019plasticity}. 
A robot $r_i$ is one hop count away from another robot $r_j$ if it lies within the communication range $d_{comm}$, i.e., $h_i = h_j + 1$ if $d(r_i, r_j) \leq d_{comm}$, where $h_i$ is the hop count of $r_i$ and $d(r_i, r_j)$ is the Euclidean distance between $r_i$ and $r_j$.
Each robot maintains its shortest hop count to the leader robot, initialized at $h_{leader} = 0$. The hop count updates are achieved through broadcasting, with each robot $r_i$ updating its hop count using $h_i = \min(h_i, h_j + 1)$.
Due to constant robot movement, a robot may temporarily move outside the communication range for a short time and then return. Its hop count is cleared only if it remains disconnected for longer than five seconds.

Robots within a hop count of up to $s$ from the leader are designated as decision-making robots, such that $S_{DM} = \{ r_i \mid h_i \leq s \}$.
To maintain the structure, each robot needs to broadcast the leader ID, its hop count, the time since its last disconnection, and the subset size parameter $s$. 

\begin{figure*}[tbp]
    \centering
    \includegraphics[width=1\linewidth]{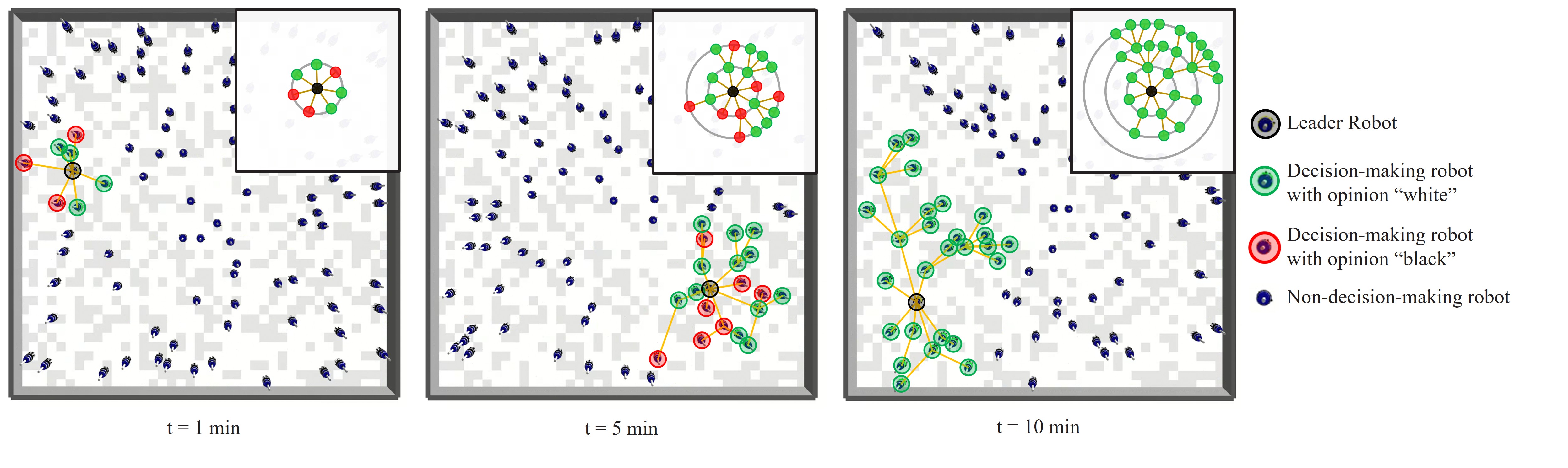}
    \caption{The collective decision-making process using a leader-based subset in an environment with black and white tiles, simulated in ARGoS. 
    The subset begins with a small group of robots holding different opinions, all directly connected to the leader (left), and expands over time until a consensus on a single opinion is reached (right). The inset graphs illustrate the formed structure at corresponding timestamps.}
    \vspace{-8pt}
    \label{fig:LCLG}
\end{figure*}

\emph{Subset Evaluation:}
The subset evaluation process is primarily managed by the leader robot, which identifies the subset size parameter ($s$) necessary to achieve a reliable final decision. The follower robots only update and pass $s$ as they receive it. 
It begins with the smallest $s$ of one to construct the decision-making subset with only the robots directly connected to the leader. The leader collects opinions from disseminating robots within its communication range, each uniquely identified by its robot ID. 
When the number of gathered opinions reaches a certain threshold $n_{op}$, indicating sufficient data collection, the leader calculates the ratio of the majority opinion collected.  
A ratio threshold $r_{op}$ is set to signal when the group is converging toward a unified opinion. To ensure the stability of this convergence, the ratio of majority opinion must remain above the threshold for a duration longer than $\tau_{op}$. If this condition holds, the leader considers the group to have reached a decision and records the opinion as the final decision associated with the current $s$. The process is repeated with increasing $s$ until $k$  consistent decisions are obtained, ensuring the reliability of the decision. We use $n_{op}=10$, $r_{op}=80\%$, $\tau_{op}=10$ seconds, and $k=2$ in our experiment.
However, if the group fails to maintain a sufficiently high majority ratio within a time limit (set as 150 seconds in our experiment), it is assumed that a decision cannot be made with the current $s$, prompting an increase in $s$ and a repetition of the process. 

\subsection{Distributed Strategy}
\emph{Subset Construction:}
In our second strategy, each robot $r_i$ in the swarm has a probability $p_i$ of being selected as a decision-making robot based on subset parameter $s_i$, where each increment of $s_i$ maps to an increase of 0.1 in $p_i$. A random value $\delta_i$ is generated independently for each robot from a uniform distribution $U(0,1)$. 
The decision-making subset is formed by the robots that satisfy $S_{DM} = \{ r_i \mid \delta_i \leq p_i \}$.
Since the selection is random, decision-making robots are not necessarily close to each other. To maintain communication between them, non-decision-making robots act as relays. Each relay can forward up to three messages, each containing a decision-making robot's ID and its opinion.

\emph{Subset Evaluation:}
Each robot in the swarm has a convergence confidence measure, denoted as $ \alpha_i $, which represents its belief about the convergence of the swarm's opinions. This confidence, ranging from 0 to 1, is updated dynamically based on local observations. Initially, each robot sets $ \alpha_i $ to 1, assuming convergence since it only knows its own opinion.
Each robot monitors the opinions of neighboring decision-making robots within its communication range $ d_{\text{comm}} $. Based on whether the observed opinion ($o_j$) matches or differs from its own ($o_i$), $ \alpha_i $ is updated accordingly to reflect the robot's evolving confidence in the swarm's convergence.
The update rule for $ \alpha_i $ is:
\begin{equation}
    \alpha_i(t+1) = 
    \begin{cases} 
        \alpha_i(t) + \gamma(1-\alpha_i(t)), & \text{if } o_i = o_j, \\ 
        \alpha_i(t) - \gamma\alpha_i(t), & \text{if } o_i \neq o_j,
    \end{cases}
\end{equation}
where $ \gamma $ is a scaling constant that determines the sensitivity of the confidence updates. A higher $\gamma$ increases the sensitivity, enabling rapid adjustments to confidence based on new observations, while a lower $\gamma$ results in more gradual confidence updates.  In our experiment, we use $\gamma=0.01$. When $o_i = o_j$, the confidence increases based on how far it is from the maximum value of 1, with smaller increments as it gets closer to full certainty. On the other hand, when $o_i \neq o_j$, the confidence decreases in proportion to its current level, meaning disagreements have a stronger effect when confidence is high. This reflects the idea that a strongly held belief in convergence should be more sensitive to conflicting opinions, encouraging the robot to reassess its confidence in the swarm's state. 
For each decrement of $\alpha_i$ by a specified step size (set to 0.05 in our experiment), the subset size parameter ($s_i$) increases by one. 
It is important to note that in the distributed strategy, $s_i$ is not a uniform value across the swarm as each robot maintains its own version of $s_i$, which may differ based on its observations and confidence levels. Consequently, the construction and self-organization of subsets are performed individually by each robot based on its own $s_i$. This means that there is no definitive, swarm-wide value for $s$ as in the leader-based implementation. Instead, the effective value is the result of the aggregation of individual robots' $s_i$ values.

\subsection{Collective Decision-Making Method}
We utilize DMVD \cite{valentini2014self} as the collective decision-making method within the selected subset of the swarm. While various collective decision-making methods could also be used, we select DMVD due to its proven ability to achieve high decision accuracy while minimizing communication bandwidth among the swarm. 
This strategy involves robots cycling through two distinct behavioral states: the exploration state and the dissemination state. In the exploration state, each robot individually assesses the quality of environmental features based on its current opinion. This assessment is conducted through sensory inputs, where each robot records the amount of time it observes the feature associated with its current opinion ($t_o$). The duration of this exploration phase ($t_e$) is sampled from an exponential distribution with a mean of $\sigma$. The quality estimate ($\hat{\rho}$) is then calculated as $\hat{\rho} = {t_o}/{t_e}$. In the dissemination state, robots broadcast their opinions to nearby robots, with the broadcast duration given by $t_d = \text{sample}(\text{Exp}(\hat{\rho} g))$,
where $g$ is a design parameter for the dissemination time. In our experiments, we set $\sigma = 10$ seconds and $g = 10$ seconds. During the dissemination phase, each robot listens to opinions from neighboring robots that are also in the process of disseminating their opinions. At the end of this state, the robot randomly adopts an opinion from its neighbors. This process establishes a positive feedback loop in which widely shared opinions are more likely to be adopted by the swarm, ultimately guiding the collective decision-making process toward a consensus. 

\begin{figure}[tbp]
    \centering
    \includegraphics[width=1\linewidth]{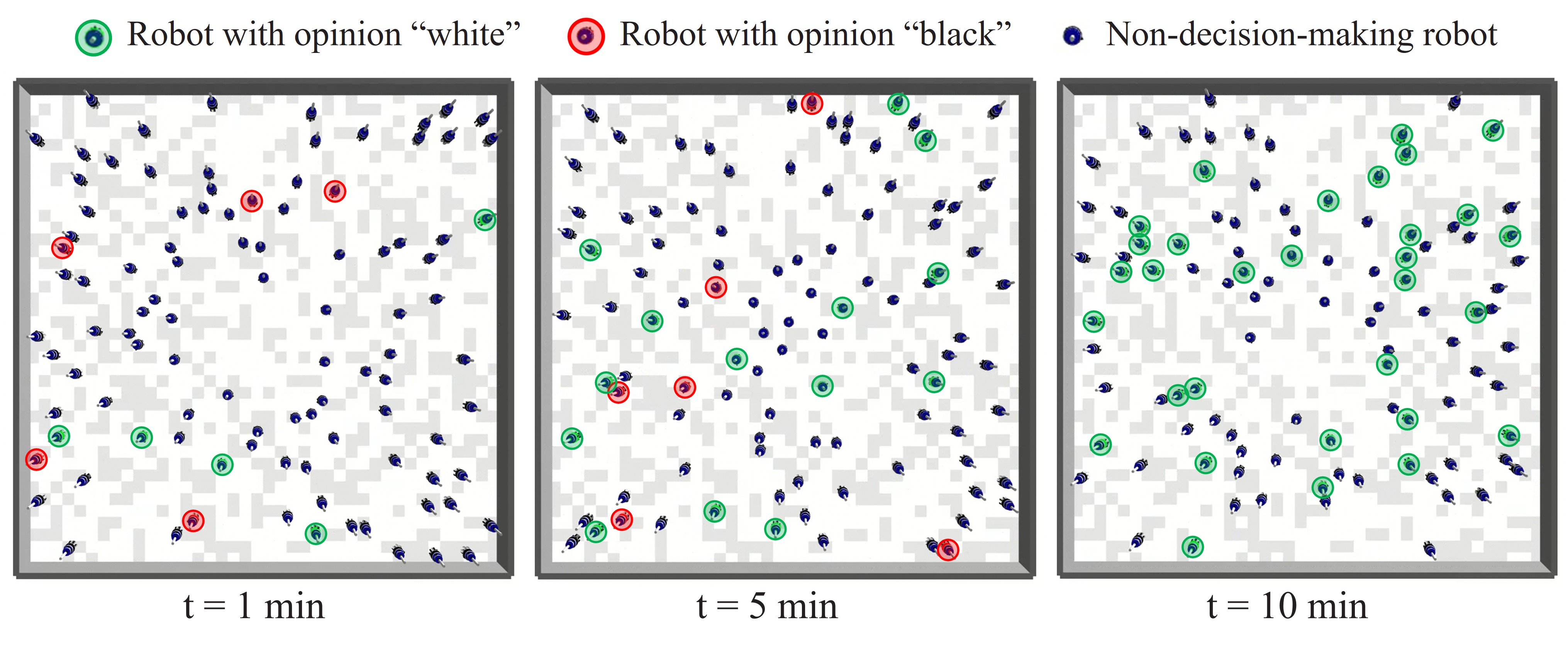}
    \caption{The collective decision-making process using a distributed subset in an environment with black and white tiles, simulated in ARGoS. 
    The subset starts with a small number of robots holding different opinions (left) and grows over time until consensus on a single opinion is reached (right).}
    \vspace{-10pt}
    \label{fig:PRCD}
\end{figure}

\subsection{Experimental Setup}

We use the ARGoS simulator \cite{pinciroli2012argos} to evaluate our method, as shown in \cref{fig:LCLG} and \cref{fig:PRCD}. It allows for scalable and large-scale testing with various swarm sizes, which would be difficult to achieve with a few physical robots. The simulations are conducted in discrete time steps, at a rate of 10 ticks per second. The environment consists of randomly distributed $20 \times 20$ cm$^2$ black and white tiles in an $8 \times 8$ m$^2$ arena. A total of 100 foot-bot robots \cite{dorigo2013swarmanoid}, representing a large robot population, are placed randomly at the start of each simulation. Each robot is assigned a random initial opinion with a 50:50 ratio, and a randomly generated ID at the beginning of each experiment. 
The robots, with a diameter of 17 cm, move using differential steering with a maximum speed of 32 cm/s and utilize proximity sensors to avoid collisions. Each robot performs a random walk, cycling between moving straight and rotating. The robot moves straight for a duration drawn randomly from an exponential distribution with a mean of 40 seconds, followed by a rotational movement lasting for a duration drawn from a uniform distribution between 0 and 4.5 seconds.
They detect black and white tiles through ground sensors and communicate with other robots using range-and-bearing communication, with a maximum range of 1 m. 
The parameters used in the experiments, mentioned in previous subsections, were optimized through trial-and-error for our setup.

\begin{figure}[tbp]
    \centering
    \includegraphics[width=1\linewidth]{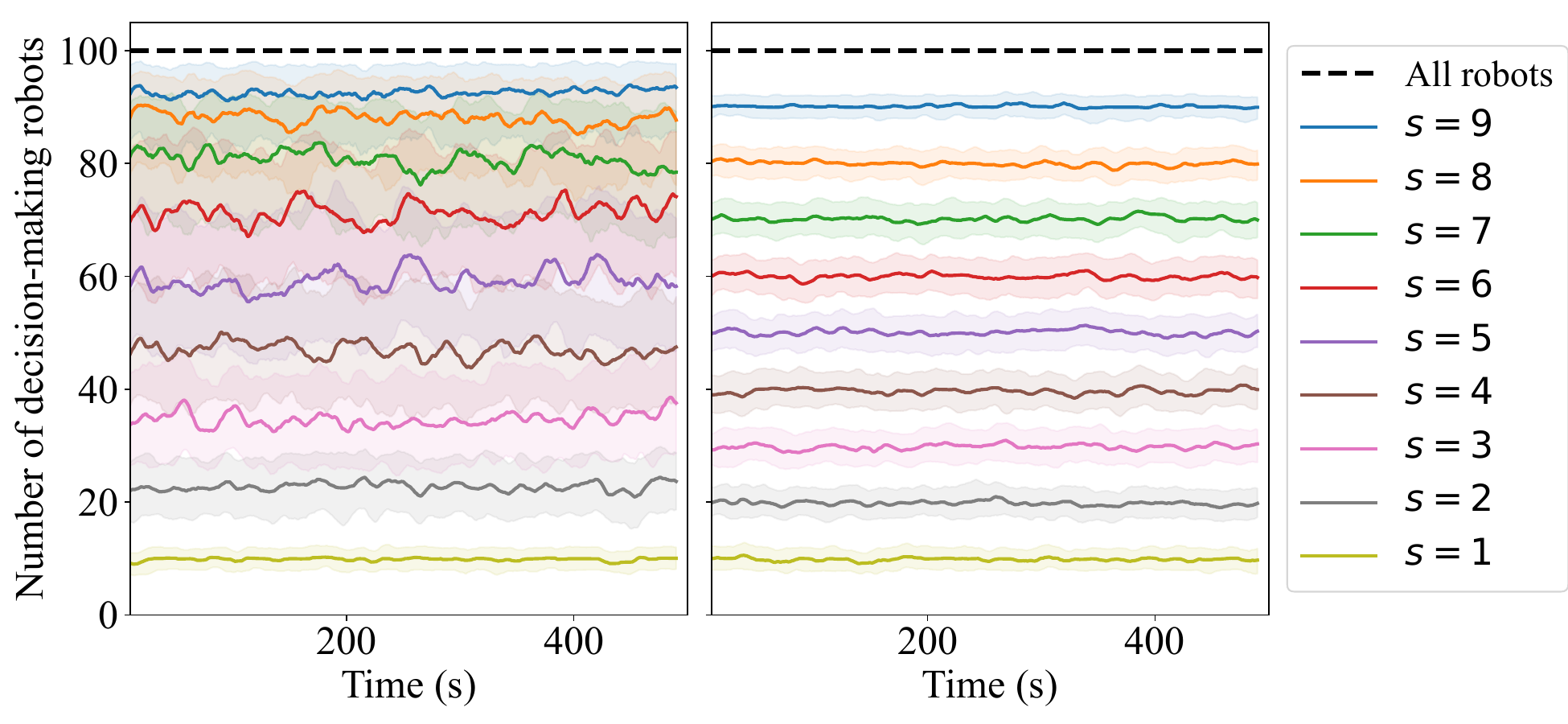}
    \caption{Number of decision-making robots over time for different subset size parameter $s$ using leader-based (left) and distributed (right) subset construction. The lines represent the median, while the shading indicates the upper and lower percentiles for the corresponding colors, derived from 100 experimental repetitions.}
    \vspace{-8pt}
    \label{fig:subset_size}
\end{figure}

\begin{figure}[tbp]
    \centering
    \includegraphics[width=1\linewidth]{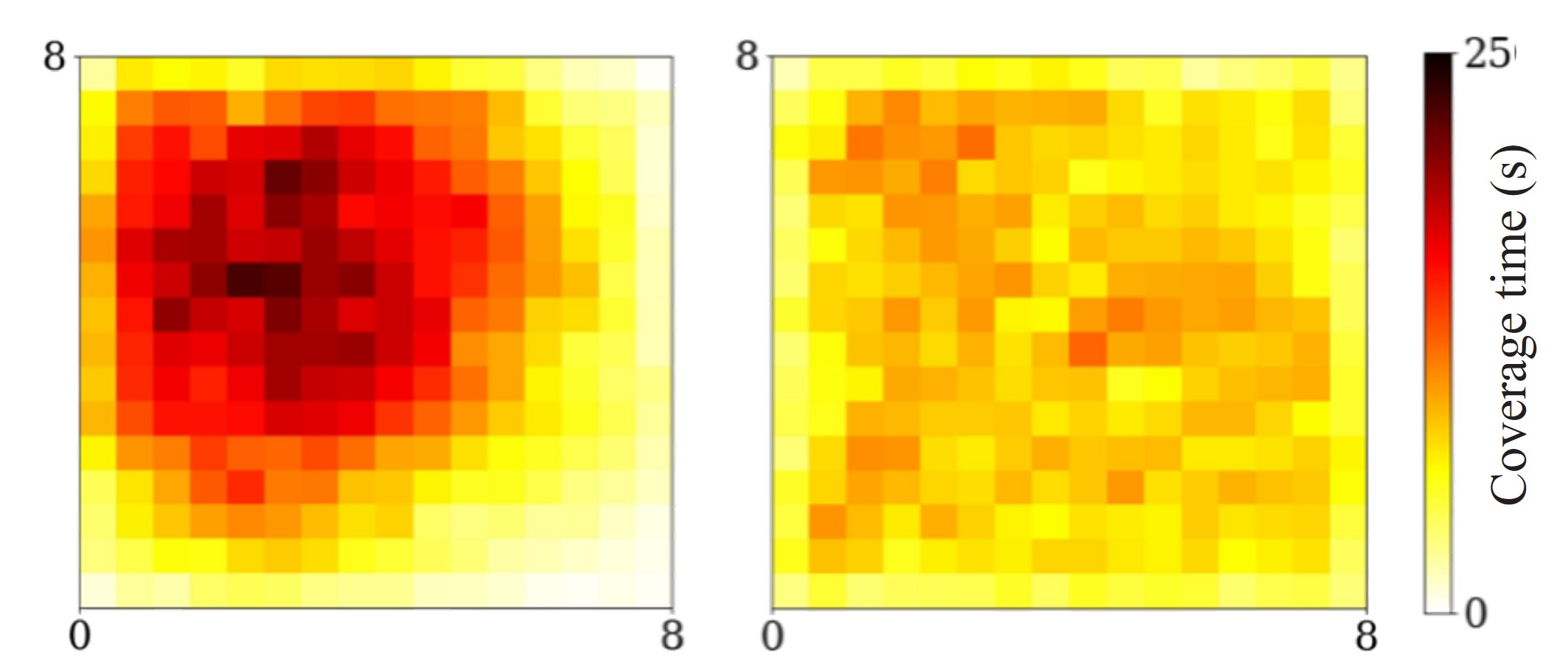}
    \caption{The coverage of decision-making robots during a 500-second runtime using leader-based (left) and distributed (right) subset construction. The color intensity represents the duration a particular location is visited by decision-making robots, with darker colors indicating longer cumulative visit times.}
    \vspace{-10pt}
    \label{fig:distribution}
\end{figure}

\section{Results}

We begin by evaluating the decision-making subset construction, examining how effectively the subsets are formed under various subset size parameters. Next, we analyze the performance of subset-based collective decision-making. Finally, we further assess the robustness of SubCDM under challenging conditions, including the presence of perception noise, reduced communication, and faulty robots.

\emph{Subset Construction:} 
The dynamics of subset construction are examined in terms of decision-making subset size (i.e. number of decision-making robots) over time. 
The behavior of the leader-based implementation in \cref{fig:subset_size} (left) stems from its recruitment mechanism, which propagates signals outward from the leader. At lower $s$, the subset remains near the leader, forming a small group. As $s$ increases, the subset expands until most reachable robots within the swarm's communication range are included ($s = 9$ in our setting). Beyond this point, the subset size plateaus, constrained by the swarm's total size (100 robots). Fluctuations occur due to robot mobility, temporarily altering the network topology, but these have little effect on overall subset size dynamics.
In the distributed implementation, the relationship between $s$ and the number of decision-making robots in the subset is linear, with the subset size increasing proportionally to $s$, as shown in \cref{fig:subset_size} (right). This relationship remains stable over time, with only minor fluctuations due to the inherent randomness in robot recruitment. 

The distribution of decision-making robots in the arena (see \cref{fig:distribution}) reveals distinct patterns of area coverage depending on the subset construction method. For the leader-based implementation, the area explored by the decision-making subset is not evenly distributed, and the robots tend to remain in certain regions of the arena for extended periods. As the leader spends more time in specific areas, possibly due to obstacles such as walls or other robots, the decision-making robots are also drawn to these areas, resulting in uneven exploration. The distributed implementation, on the other hand, demonstrates relatively even coverage of the arena. Since each robot has a set probability of becoming a decision-making robot, the recruitment process is random, and over time, this randomness leads to a roughly uniform distribution of robots. We also calculate Moran's Index \cite{moran_notes_1950} to measure spatial autocorrelation that quantifies how similar the spatial distribution of decision-making robots is to their neighbors, over one hundred runs. The leader-based implementation yields a value of 0.875, while the distributed implementation yields 0.366, indicating stronger spatial clustering in the distribution of decision-making robots for the leader-based implementation. This difference in spatial distribution can also be seen in the heatmaps presented in \cref{fig:distribution}.

\begin{figure}[tbp]
    \centering
    \includegraphics[width=0.8\linewidth]{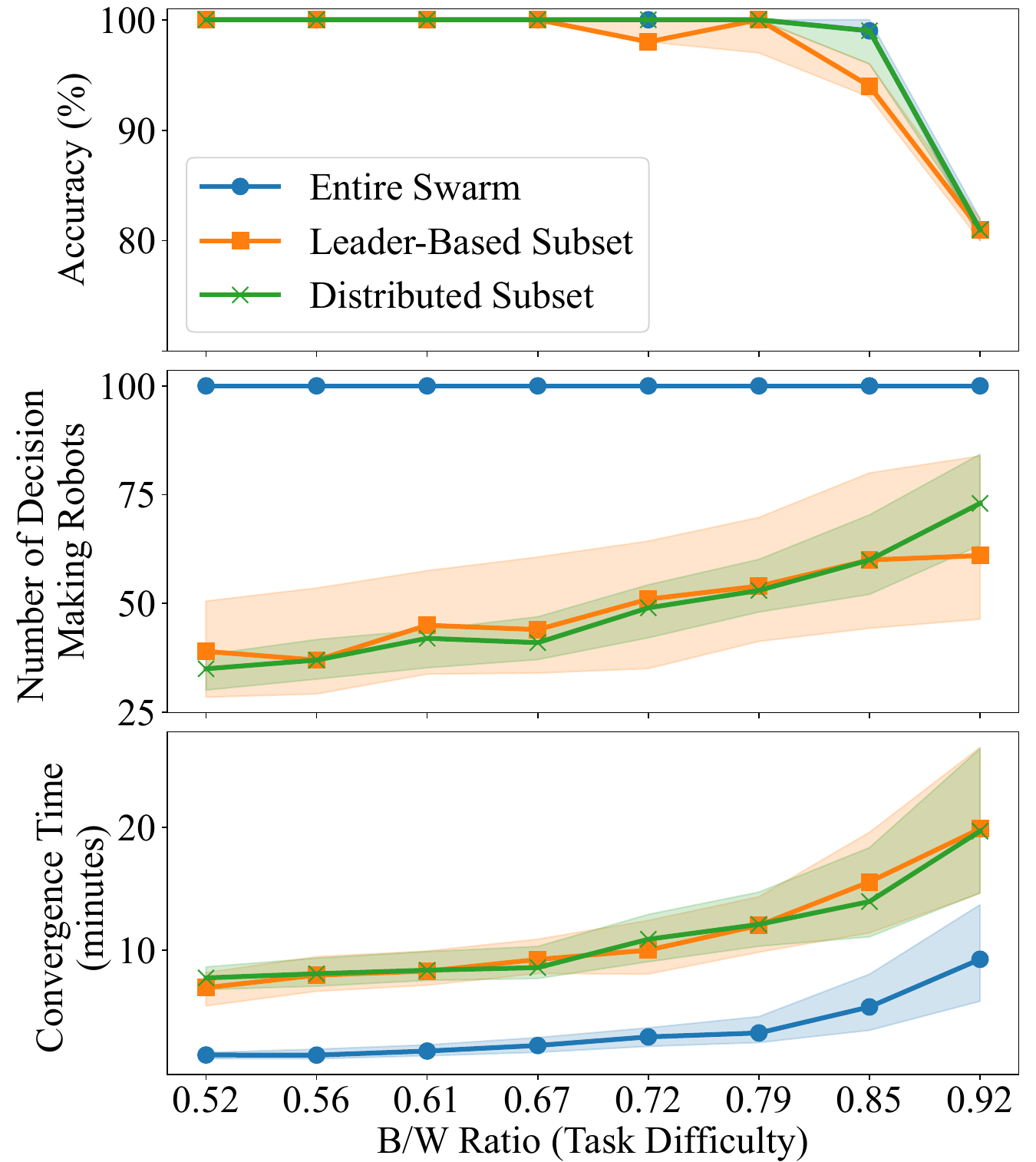}
    \caption{Accuracy of decision (top), number of decision-making robots in steady state (middle), and convergence time (bottom) with varying task difficulty. The lines represent the median, while the shading indicates the upper and lower percentiles for the corresponding colors, derived from 100 experimental repetitions.}
    \vspace{-20pt}
    \label{fig:main}
\end{figure}

\emph{Subset-Based Collective Decision-Making: }
As shown in \cref{fig:LCLG}, the leader-based subset starts with a small group of robots directly connected to the leader and grows over time as more robots, progressively farther from the leader, are included until consensus is reached. The distributed subset, shown in \cref{fig:PRCD}, also grows until consensus is reached, but with decision-making robots more spread out.

To evaluate the system performance under varying conditions, we vary the decision-making task difficulty by adjusting the proportion of black and white tiles in the arena. Specifically, the proportion of black tiles is adjusted in increments of 2\%, ranging from 34\% to 48\%, resulting in black-to-white ratios between 0.52 and 0.92. As the black-to-white ratio approaches 1, the decision-making task becomes more difficult as it becomes harder for the swarm to identify the dominant color. Each scenario is repeated 100 times. 
\cref{fig:main} (top) shows the accuracy of our method across decision-making task difficulties. As a comparison, we also conduct collective decision-making experiments with the entire swarm using DMVD across different decision-making task difficulties. We consider a final decision to be made if a large portion of robots steadily hold one opinion. Specifically, we use a threshold of at least 80\% of robots holding the same opinion for at least 30 seconds. A general trend is that accuracy decreases as task difficulty increases. Both implementations of SubCDM maintain accuracy similar to that of the entire swarm across varying task difficulties. 
The number of decision-making robots at steady state is shown in \cref{fig:main} (center), where higher task difficulty (corresponding to a black-to-white ratio closer to 1) leads to recruiting a larger number of decision-making robots. In contrast to the entire swarm, which consistently uses all 100 robots, our approach employs fewer robots, with the subset size varying based on task difficulty. As task difficulty increases, convergence times also lengthen (see \cref{fig:main} (bottom)), as the swarm requires more time to resolve the greater environmental complexity. While using the entire swarm results in faster convergence by involving all robots, our subset-based approach is beneficial for broader applications, including those involving concurrent task execution.

\begin{figure}[tbp]
    \centering
    \includegraphics[width=1\linewidth]{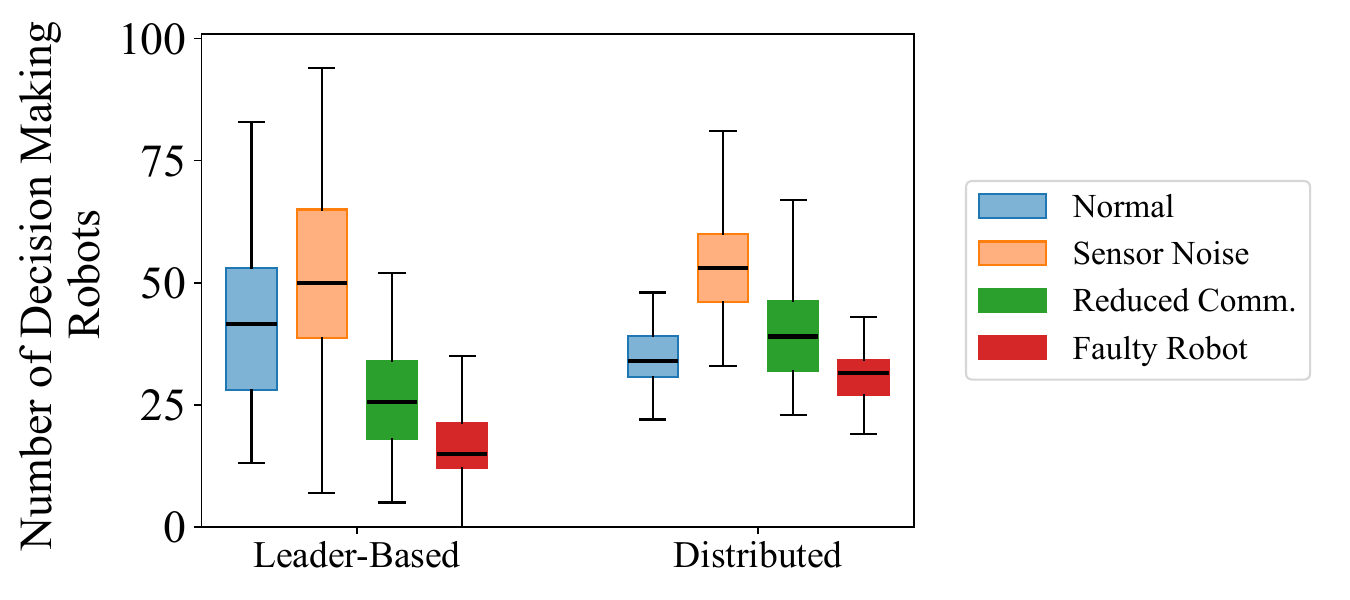}
    \caption{Number of decision-making robots under different disturbance types across 100 runs, with the black horizontal lines indicating the medians.}
    \vspace{-21pt}
    \label{fig:robust}
\end{figure}

\emph{System Robustness: }
We evaluate the robustness of our method under perception noise, reduced communication, and faulty robots (see \cref{fig:robust}). We integrate distributed leader election \cite{vasudevan2004design} into our leader-based implementation. Perception noise, where some sensor readings are incorrect, disrupts the decision-making process by reducing the reliability of the opinions formed by decision-making robots, leading to slower convergence as more time is needed to reach a consensus. Both implementations of our approach adapt to perception noise in the same way that they adapt to decision-making task difficulty, by dynamically growing the size of the decision-making subset. Essentially, the effect of perception noise is similar to increasing task difficulty, as both increase the complexity of the environment and challenge the swarm's ability to identify the dominant color. Higher levels of perception noise are compensated by including more robots in the decision-making subset. This adaptivity enables the swarm to maintain robustness in scenarios with low or moderate noise levels. 
In scenarios with reduced communication, the frequency of information exchange between robots is significantly lowered and the decision-making process slows, leading to the inclusion of more robots in the decision-making subset. Reduced communication also disrupts the hierarchical structure, which can significantly decrease the subset size if it depends on the structure, as in the leader-based implementation. As a result, the overall number of decision-making robots in the leader-based implementation is reduced, and decision-making quality decreases. In contrast, the distributed implementation, which does not rely on any structure, is able to compensate for slower decision-making by including more robots in the decision-making subset. 
To investigate the effect of faulty robots, we simulate a scenario where robots experience intermittent failures, such as communication or mobility disruptions. With a 10\% probability, robots malfunction for a brief period before resuming normal operation. This malfunction can affect all robots, whether decision-making or not. When a robot malfunctions, it ceases communication and movement, simulating real-world failures. The results show that the leader-based implementation is sensitive to these faults, especially when the leader malfunctions, which significantly affects the number of decision-making robots. The distributed leader election mechanism allows another robot to take over the leader role; however, this requires a reassessment of the situation, leading to a noticeable decrease in the number of robots involved in decision-making. In contrast, the distributed implementation demonstrates greater robustness, with an 11\% reduction in number of decision-making robots, closely matching the faulty rate, indicating lower dependency on specific robots.

\section{Conclusion}
We presented SubCDM, a mechanism to autonomously construct a subset of robots in swarms to perform collective decision-making, addressing the challenge of reducing the number of active robots while maintaining decision-making performance. This mechanism consists of two components: subset construction and subset evaluation. 
Subset construction allows the robots to self-organize into a decision-making subset, while subset evaluation scales its size based on task difficulty.
Together with a collective decision-making strategy, these components enable the swarm to efficiently allocate active decision-making robots, thereby freeing the remaining robots for other decision-making tasks or preserving their energy. 
We evaluated our method with two decentralized implementations, leader-based and distributed, both relying solely on local information. They achieve accuracy comparable to full-swarm participation while reducing the number of active decision-making robots. The leader-based strategy forms a spatially clustered subset around a leader, facilitating efficient communication but sensitive to disruptions that affect the hierarchical structure, such as reduced connectivity or faulty robots. In contrast, the distributed strategy is more resilient to such challenges but relies on non-decision-making robots as information relays, as decision-making robots are not necessarily adjacent.
For future work, we aim to explore a resource-efficient approach in scenarios where the distribution of features is not uniform. We intend to investigate how our method can be adapted to handle situations with varying densities and distributions of features, which is more challenging due to the increased complexity in achieving accurate decision-making.

\bibliography{IROS25ref_short}
\bibliographystyle{ieeetr}



\end{document}